\documentclass{article}
\usepackage{spconf,amsmath,graphicx}
\usepackage{graphicx}
\usepackage{amsmath,amssymb,amsfonts}
\usepackage{multirow}
\usepackage[shortlabels]{enumitem}
\usepackage[cal=cm]{mathalfa}
\usepackage{subfigure}
\usepackage[algoruled,boxed,lined,ruled]{algorithm2e}
\usepackage{tikz}
\usepackage{pgfplots}
\pgfplotsset{compat=1.14}

\title{Knowledge Distillation for Optimization of \\Quantized Deep Neural Networks}
\name{Sungho Shin, Yoonho Boo, and Wonyong Sung\thanks{Thanks to XYZ agency for funding.}}
\address{Department of Electrical and Computer Engineering\\
Seoul National University\\
Seoul, 08826 Korea\\
\texttt{sungho.develop@gmail.com, wysung@snu.ac.kr}}
\definecolor{applegreen}{rgb}{0.7, 0.93, 0.36}

\begin{document}
\ninept
\maketitle
%
\begin{abstract}
Knowledge distillation (KD) technique that utilizes a pre-trained teacher model for training a student network is exploited for the optimization of quantized deep neural networks (QDNNs). We considered the choice of the teacher network and also investigate the effect of hyperparameters for KD. We have tried several large floating-point models and quantized ones as the teacher. The experiments show that the softmax distribution produced by the teacher network is more important than its performance for effective KD training. Since the softmax distribution of the teacher network can be controlled by KD’s hyperparameters, we analyze the interrelationship of each KD component for quantized DNN training. We show that even a small teacher model can achieve the same distillation performance as a larger teacher model. We also propose the \textit{gradual soft loss reducing} (GSLR) technique for robust KD based QDNN optimization, which controls the mixing ratio of hard and soft losses during training.   
\end{abstract}
\begin{keywords}
Deep neural network, quantization, knowledge distillation, fixed-point optimization
\end{keywords}
\section{Introduction}
\label{sec_intro}
Deep neural networks (DNNs) usually require a large number of parameters, thus it is very necessary to reduce the size of the model to operate it in embedded systems. Quantization is a widely used compression technique, and even 1- or 2-bit models can show quite good performance. However, it is necessary to train the model very carefully not to lose the performance when only low-precision arithmetic is allowed. Many QDNN papers have suggested various types of quantizers or complex training algorithms~\cite{hubara2017quantized,hwang2014fixed,xu2018alternating,zhou2016dorefa,zhou2017balanced}.

Knowledge distillation (KD) that trains small networks using larger networks for better performance~\cite{hinton2015distilling,bucilua2006model}. KD employs the soft-label generated by the teacher network to train the student network. Leveraging the knowledge contained in previously trained networks has attracted attention in many applications for model compression~\cite{tang2016recurrent,song2018neural,asami2017domain,wang2019deepvid} and learning algorithms~\cite{romero2014fitnets,kulkarni2017knowledge,park2019relational,yim2017gift}. Recently, the use of KD for the training of QDNN has been studied~\cite{mishra2018apprentice,polino2018model}. However, there are many design choices to explore when applying KD to QDNN training. The work in~\cite{mishra2018apprentice} studied the effects of simultaneous training or pre-training in teach model design. The result is rather expected; employing a pre-trained teacher model is advantageous when considering the performance of the student network. However,~\cite{mirzadeh2019improved} mentioned that a too large teacher network does not help improve the performance of the student model.

In this work, we exploit KD with various types of teacher networks that include full-precision~\cite{mishra2018apprentice,polino2018model} model, quantized one, and teacher-assistant based one~\cite{mirzadeh2019improved}. The analysis results indicate that, rather than the type of the model, the distribution of the soft label is critical to the performance improvement of the student network. Since the distribution of the soft label can be controlled by the \textit{temperature} and the \textit{size of the teacher network}, we try to show how well-selected \textit{temperature} can improve the QDNN performance dramatically even with a small teacher network. Further, we suggest a simple KD training scheme that adjusts the mixing ratio of hard and soft losses during training for obtaining stable performance improvements. We name it as the \textit{gradual soft loss reducing} (GSLR) technique. GSLR employs both soft and hard losses equally at the beginning of the training, and gradually reduces the ratio of the soft loss as the training progresses. 

This paper is organized as follows. Section~\ref{sec_QDNNKD} describes how the QDNN can be trained with KD and explains why the hyperparameters of KD are important. Section~\ref{sec_experiment} shows the experimental results and we conclude the paper in Section~\ref{sec_conclude}.

\section{Quntized Deep Neural Network Training Using Knowledge Distillation}
\label{sec_QDNNKD}
In this section, we first briefly describe the conventional neural network quantization method and also depict how QDNN training can be combined with KD. We also explain the hyperparameters of KD and their role in QDNN training.

\subsection{Quantization of deep neural networks and knowledge distillation}
\label{secsec_revisit}
The deep neural network parameter vector, $\mathbf{w}$, can be expressed in $2^b$ level when quantized in $b$-bit. Since we usually use a symmetric quantizer, the quantized weight vector $\text{Q}(\mathbf{w})$ can be represented using \eqref{eq_1bit_quant} or \eqref{eq_2bit_quant} for the case of $b = 1$ or $b > 1$ as follows:
\begin{align}
\text{Q}^{1}(\mathbf{w})&=\text{Binarize}(\mathbf{w}) = \Delta\cdot\text{sign}(\mathbf{w})\label{eq_1bit_quant}\\
\text{Q}^{b}(\mathbf{w}) &= \text{sign}(\mathbf{w})\cdot\Delta\cdot\text{min}\Big\{\Big\lfloor\Big(\frac{|\mathbf{w}|}{\Delta}+0.5\Big)\Big\rfloor, \frac{(M-1)}{2}\Big\}\label{eq_2bit_quant}
\end{align}
where $M$ is the number of quantization levels ($2^b-1$) and $\Delta$ represents the quantization step size. $\Delta$ can be computed by L2-error minimization between floating and fixed-point weights or by the standard deviation of the weight vector~\cite{hwang2014fixed,rastegari2016xnor,shin2017fixed}.

Severe quantization such as 1- or 2-bit frequently incurs large performance degradation. Retraining technique is widely used to minimize the performance loss \cite{sung2015resiliency}. When retraining the student network, forward, backward, and gradient computations should be conducted using quantized weights but the computed gradients must be added to full-precision weights~\cite{hubara2017quantized,hwang2014fixed,xu2018alternating,zhou2016dorefa,zhou2017balanced}.

The probability computation in deep neural networks usually employ the softmax layer.  Logit, $\mathbf{z}$, is fed into the softmax layer and generates the probability of each class, $\mathbf{p}$, using $p_i = \frac{exp(z_i/\tau)}{\sum_{j}exp(z_j/\tau)}$. $\tau$ is a hyperparameter of KD known as the \textit{temperature}.  A high value of $\tau$ softens the probability distribution. KD employs the probability generated by the teacher network as a soft label to train the student network, and the following loss function is minimized during training. 
\begin{align}
\mathcal{L}(\mathbf{w}_{\text{S}}) = (1-\lambda)\mathcal{H}(y,p^{\text{S}}) + \lambda\mathcal{H}(p^{\text{T}},p^{\text{S}})\label{eq_KD_loss}
\end{align}
\noindent where $\mathcal{H}(\cdot)$ denotes a loss function, $y$ is the ground truth hard label, $\mathbf{w}_\text{S}$ is the weight vector of the student network, $p^\text{T}$ and $p^\text{S}$ are the probabilities of the teacher and student networks, and $\lambda$ is a \textit{loss weighting factor} for adjusting the ratio of soft and hard losses. 

A recent paper~\cite{mirzadeh2019improved} suggests the teacher, teacher-assistant, and student models because the effect of KD gradually decreases when the size difference between the teacher and student networks becomes too large. This performance degradation is due to the capacity limitation of the student model. Since QDNN limits the representation level of the weight parameters, the capacity of a quantized network is reduced when compared with the full-precision model. Therefore, QDNN training with KD is more sensitive to the \textit{size of the teacher network}. We consider the optimization of three hyperparameters described above (\textit{temperature}, \textit{loss weighting factor}, and \textit{size of the teacher network}). Algorithm~\ref{algo} describes how to train QDNN with KD.

\begin{algorithm}[t]
\caption{QDNN training with KD}
\label{algo}
\SetKwInOut{Input}{Input}
\SetKwInOut{Output}{Output}
\SetKwInOut{Initialization}{Initialization}
\SetKw{KwReturn}{Return}
\SetKw{KwAnd}{and}

\Initialization{$\mathbf{w}_\text{T}$: Pretrained teacher model,\\
$\mathbf{w}_\text{S}$: Pretrained student model,\\
$\lambda$: Loss wegithed factor, $\tau$: Temperature}

\Output{$\mathbf{w}_{\text{S}}^{\text{q}}$: Quantized student model}

\While{not converged} {
$\mathbf{w}_{\text{S}}^{\text{q}} = \text{Quant}(\mathbf{w}_{\text{S}})$ 

\text{Run forward teacher ($\mathbf{w}_\text{T}$) and student model ($\mathbf{w}_{\text{S}}^{\text{q}}$)}

\text{Compute distillation loss} $\mathcal{L}(\mathbf{w}_{\text{S}}^{\text{q}},\, \lambda)$

\text{Run backward and compute gradients $\frac{\partial\mathcal{L}(\mathbf{w}_{\text{S}}^{\text{q}})}{\partial\mathbf{w}_{\text{S}}^{\text{q}}}$}

$ \mathbf{w}_{\text{S}} = \mathbf{w}_{\text{S}} - \eta \cdot \nabla \frac{\partial\mathcal{L}(\mathbf{w}_{\text{S}}^{\text{q}})}{\partial\mathbf{w}_{\text{S}}^{\text{q}}}$;

}

\KwReturn{$\mathbf{w}_{\text{S}}^{\text{q}}$}
\label{alg:HLHL}
\end{algorithm}

\subsection{Teacher model selection for KD}
\label{secsec_qdnn_training}
In this section, we try to find the best teacher model for QDNN training with KD. We consider three different approaches. The first one is training the full-precision teacher and student networks independently and applies KD when fine-tuning the quantized student model as suggested in~\cite{mishra2018apprentice,polino2018model}. The second is training a medium-sized teacher assistant network with a very large teacher model, and then optimizing the student network using the teacher assistant model as suggested by~\cite{mirzadeh2019improved}. The last approach is using a quantized teacher model with the possibility of the student learning something on quantization.
\begin{table}[t]
\centering
\caption{Train and test accuracies of the quantized ResNet20 that trained with various KD methods on CIFAR-10 dataset. `$\text{T}_\text{L}$', `T', `S', `(F)', and `(Q)' denote large teacher, teacher, student, (full-precision), and (quantized), respectively. HD is a conventional training using hard loss. $\tau$ represents the \textit{temperature}. Note that all the student networks are 2-bit QDNN and the results are the average of five times running.}
 \vspace{-0.3cm}
\label{table:macroscopic}
\begin{tabular}{cccc}
\\
\hline\hline
Method & Status & Train Acc. & Test Acc. \\ \hline
\multirow{3}{*}{T(F)-S} & T (float) & 99.95 & 94.02 \\ 
 & S ($\tau$ = 5) & 98.02 & 92.25 \\ 
 & S ($\tau$ = 2) & 98.76 & 91.94 \\ \hline
\multirow{3}{*}{$\text{T}_\text{L}$(F)-T(F)-S} & $\text{T}_\text{L}$ (float) & 99.99 & 95.24 \\ 
 & T (float) & 99.99 & 94.8 \\ 
 & S ($\tau$ = 10)& 97.78 & 92.18 \\ \hline
\multirow{3}{*}{$\text{T}_\text{L}$(F)-T(Q)-S} & $\text{T}_\text{L}$ (float) & 99.99 & 95.24 \\ 
 & T (4-bit) & 99.99& 94.46\\ 
 & S ($\tau$ = 10)& 97.79 & 92.14 \\ \hline
\multirow{2}{*}{T(Q)-S} & T (8-bit) & 99.99 & 94.34 \\ 
 & S ($\tau$ = 10)& 97.53 & 92.02 \\ \hline
HD & Conventional & 98.91 & 91.71 \\ \hline\hline
\end{tabular}
\end{table}
\begin{figure}[t]
    \centering
    \includegraphics[width=0.7\linewidth]{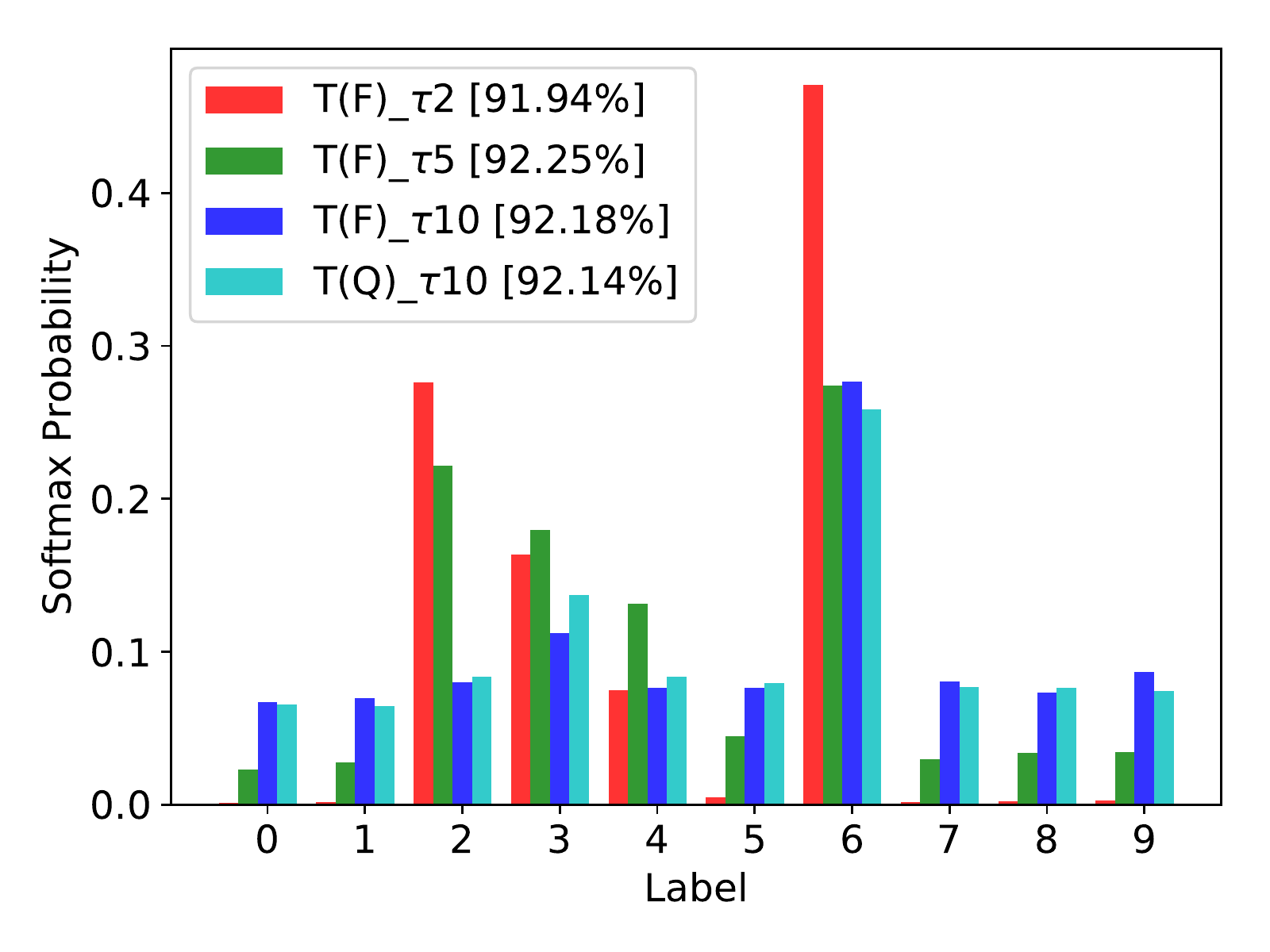}
    \vspace{-0.5cm}
    \caption{Example of the softmax distribution for label 6 from the teacher models in \tablename~\ref{table:macroscopic}. The numbers in square brackets are the CIFAR-10 test accuracies of the student networks that trained by each teacher model.}
    \label{fig:softmax_distribute}
\end{figure}

\tablename~\ref{table:macroscopic} compares the results of these three approaches. \figurename~\ref{fig:softmax_distribute} also shows the softmax distributions when the teacher models and temperatures vary. 
The test accuracy of the quantized ResNet20 trained using hard loss was 91.71\%. The results using various KD approaches indicate the following information. First, whether the teacher network is quantized or not, the performance of the student network is not much different. Secondly, employing the teacher assistant network~\cite{mirzadeh2019improved} does not help increase the performance. The performance is similar to that of a general KD. Thirdly, KD training with full-precision teacher network~\cite{mishra2018apprentice,polino2018model} is significantly better than conventional training when $\tau$ is 5. But, no performance increase is observed when $\tau$ is 2. Lastly, the teacher models that achieve the student network accuracy of 92.14\%, 92.18\%, and 92.25\% have a similar softmax distribution. However, T(F)-S with $\tau=2$ shows a quite sharp softmax shape, and the resulting performance is similar to that of hard-target training. 

These points indicate that the softmax distribution is the key to lead effective KD training. Although the different teacher models generate dissimilar softmax distributions, we can control the shapes by using the \textit{temperature}. A detailed discussion about hyperparameters is provided in the following subsections.

\subsection{Discussion on hyperparameters of KD}
\label{secsec_discussion_HP}
As we mentioned in Section~\ref{secsec_revisit} and~\ref{secsec_qdnn_training}, the hyperparameters \textit{temperature} ($\tau$), \textit{loss weighting factor} ($\lambda$), and \textit{size of teacher network} ($N$) can significantly affect the QDNN performance. Previous works usually fixed these hyperparameters when training QDNN with KD. For example, \cite{mishra2018apprentice} always fixes $\tau$ to 1, and \cite{polino2018model} holds it to 1 or 5 depending on the dataset. However, these three parameters are closely interrelated. For example, \cite{mirzadeh2019improved} points out that when the teacher model is very large compared to the student model, the softmax information produced by the teacher network become sharper, making it difficult to transfer the knowledge of the teacher network to the student model. However, even in this case, controlling the \textit{temperature} may be able to make it possible. Therefore, when the value of one hyperparameter is changed, the others also need to be adjusted carefully. Thus, we empirically analyze the effect of KD's hyperparameters. In addition, we introduce the \textit{gradual soft loss reducing} (GSLR) technique that aids to improve the performance of QDNN dramatically. The GSLR is a KD training method that gradually increases the reflection ratio of the hard loss.



\begin{table}[t]
\centering
\caption{Train and test accuracies (\%) of the teacher networks on the CIFAR-10 and the CIFAR-100 datasets. `WRN20x$N$' denotes WideResNet with a wide factor of `$N$'.}
\label{table:teachers_cifar10}
 \vspace{-0.3cm}
\begin{tabular}{cccccc}
\\\hline\hline
CIFAR-10 & Train & Test & CIFAR-100 & Train & Test \\ \hline
ResNet20 & 99.62 & 92.63 & ResNet20 & 90.12 & 68.43 \\ 
WRN20x1.2 & 99.83 & 92.93 & WRN20x1.2 & 94.92 & 69.64 \\ 
WRN20x1.5 & 99.93 & 93.48 & WRN20x1.5 & 98.63 & 71.80 \\ 
WRN20x1.7 & 99.95 & 94.02 & WRN20x1.7 & 99.36 & 72.17 \\ 
WRN20x2 & 99.95 & 94.36 & WRN20x2 & 99.82 & 74.03 \\ 
WRN20x5 & 100 & 95.24 & WRN20x3 & 99.95 & 76.31 \\ 
WRN20x10 & 100 & 95.23 & WRN20x4 & 99.95 & 77.93 \\ 
 &  &  & WRN20x5 & 99.98 & 78.17 \\ 
 &  &  & WRN20x10 & 99.98 & 78.68 \\ \hline\hline
\end{tabular}
\end{table}
\begin{table}[t]
\centering
\caption{Training results of full-precision and 2-bit quantized ResNet20 on CIFAR-10 and CIFAR-100 datasets in terms of accuracy (\%). The models are trained with hard loss only.}
\label{table:underover}
 \vspace{-0.3cm}
\begin{tabular}{cccc}
\\\hline\hline
\multicolumn{2}{c}{} & Train acc. & Test acc. \\ \hline
\multirow{2}{*}{CIFAR-10} & Full-precision & 99.62 & 92.63 \\ \cline{2-4} 
 & 2-bit quantized & 98.92 & 91.71 \\ \hline\hline
\multirow{2}{*}{CIFAR-100} & Full-precision & 90.12 & 68.43 \\ \cline{2-4} 
 & 2-bit quantized & 77.61 & 65.23 \\ \hline\hline
\end{tabular}
\end{table}
\section{Experimental Results}
\label{sec_experiment}
\subsection{Experimental setup}
\label{secsec_expsetup}
\textbf{Dataset:} We employ CIFAR-10 and CIFAR-100 datasets for experiments. CIFAR-10 and CIFAR-100 consist of 10 and 100 classes, respectively. Both datasets contain 50K training images and 10K testing images. The size of each image is 32x32 with RGB channels.

\noindent\textbf{Model configuration \& training hyperparameter:} To analyze the impact of hyperparameters of KD on QDNN training, we train WideResnet20x$N$ (WRN20x$N$)~\cite{zagoruyko2016wide} as the teacher networks, where $N$ is set to 1, 1.2, 1.5, 1.7, 2, 3, 4, 5, and 10. When $N$ is 1, the network structure is the same with ResNet20~\cite{he2016deep}. All the train and the test accuracies of the teacher networks on CIFAR-10 and CIFAR-100 datasets are reported in~\tablename~\ref{table:teachers_cifar10}. We employ ResNet20 as the student network for both the CIFAR-10 and CIFAR-100 datasets. If the network size is large enough considering the size of the dataset, which means over-parameterized, most quantization method works well~\cite{sung2015resiliency}. Therefore, to evaluate a quantization algorithm, we need to employ a small network that is located in the under-parameterized region~\cite{sung2015resiliency,boo2019memorization}. Although the full-precision ResNet20 model is over-parameterized, which means near 100\% training accuracy, the 2-bit network becomes under-parameterized on the CIFAR-10 dataset. Likewise, on the CIFAR-100 dataset, both the full-precision and the quantized models are under-parameterized. Thus,  it is a good network configuration to evaluate the effect of KD on QDNN training. We report the train and the test accuracies for ResNet20 on CIFAR-10 and CIFAR-100 in \tablename~\ref{table:underover}.

\begin{table}[t]
\centering
\caption{Results of QDNN training with KD on ResNet-20 for CIFAR-10 and CIFAR-100 dataset. `WRN',`RN', `SM', `DS' represent WideResNet, ResNet, student model, and deeper student, respectively.}
\label{table:compare}
 \vspace{-0.3cm}
\setlength\tabcolsep{1.2pt}
\begin{tabular}{ccccccc}
\\\hline\hline
CIFAR10 & \multicolumn{2}{c}{Teacher (full-precision)} & \multicolumn{4}{c}{Student (2-bit)} \\ \hline
 & \begin{tabular}[c]{@{}c@{}}\# params (M)\\ (model name)\end{tabular} & \begin{tabular}[c]{@{}c@{}}Test\\ (\%)\end{tabular} & \begin{tabular}[c]{@{}c@{}}\# params (M)\\ (model name)\end{tabular}& \begin{tabular}[c]{@{}c@{}}Test\\ (\%)\end{tabular} & $\tau$ & $\lambda$ \\ \hline\hline
\multirow{3}[8]{*}{QDistill} & \multirow{2}[4]{*}{\begin{tabular}[c]{@{}c@{}}5.3 \\ (small network)\end{tabular}} & \multirow{2}[4]{*}{89.7} & \begin{tabular}[c]{@{}c@{}}0.3\\ (SM 2)\end{tabular} & 74.2 & 5 & 0.5 \\ \cline{4-7} 
 &  &  & \begin{tabular}[c]{@{}c@{}}5.8\\ (DS)\end{tabular} & 89.3 & 5 & 0.5 \\ \cline{2-7} 
 & \begin{tabular}[c]{@{}c@{}}145 \\ (WRN28x20)\end{tabular} & 95.7 & \begin{tabular}[c]{@{}c@{}}82.7\\ (WRN22x16)\end{tabular} & 94.23 & 5 & 0.5 \\ \hline
\multirow{2}[3]{*}{Apprentice} & \begin{tabular}[c]{@{}c@{}}0.66\\ (RN44)\end{tabular} & 93.8& \begin{tabular}[c]{@{}c@{}}0.27\\ (RN20)\end{tabular} & 91.6 & 1 & 0.5 \\ \cline{2-7} 
 & \begin{tabular}[c]{@{}c@{}}0.66\\ (RN44)\end{tabular} & 93.8 & \begin{tabular}[c]{@{}c@{}}0.47\\ (RN32)\end{tabular} & 92.6 & 1 & 0.5 \\ \hline
\multirow{2}[4]{*}{Ours} & \begin{tabular}[c]{@{}c@{}}0.61\\ (WRN20x1.5)\end{tabular} & 93.5 & \begin{tabular}[c]{@{}c@{}}0.27\\ (RN20)\end{tabular} & 92.52 & 10 & 0.5 \\ \cline{2-7} 
 & \begin{tabular}[c]{@{}c@{}}0.38\\ (WRN20x1.2)\end{tabular} & 92.9 & \begin{tabular}[c]{@{}c@{}}0.27\\ (RN20) 1-bit\end{tabular} & 91.3 & 3 & 0.5 \\ \hline\hline
CIFAR100 & \multicolumn{2}{c}{Teacher (full-precision)} & \multicolumn{4}{c}{Student (2-bit)} \\ \hline
 & \begin{tabular}[c]{@{}c@{}}\# params (M)\\ (model name)\end{tabular} & \begin{tabular}[c]{@{}c@{}}Test\\ (\%)\end{tabular} &\begin{tabular}[c]{@{}c@{}}\# params (M)\\ (model name)\end{tabular} & \begin{tabular}[c]{@{}c@{}}Test\\ (\%)\end{tabular} & $\tau$ & $\lambda$ \\ \hline\hline
QDistill & \begin{tabular}[c]{@{}c@{}}36.5\\ (WRN28x10)\end{tabular} & 77.2 & \begin{tabular}[c]{@{}c@{}}17.2\\ (WRN22x8)\end{tabular} & 49.3 & 5 & 0.5 \\ \hline
Guided & \begin{tabular}[c]{@{}c@{}}22.0\\ (AlexNet)\end{tabular} & 65.4 & \begin{tabular}[c]{@{}c@{}}22.0\\ (AlexNet)\end{tabular} & 64.6 & - & - \\ \hline
\multirow{2}[4]{*}{Ours} & \begin{tabular}[c]{@{}c@{}}0.39\\ (WRN20x1.2)\end{tabular} & 69.64 & \begin{tabular}[c]{@{}c@{}}0.28\\ (RN20)\end{tabular} & 66.6 & 2 & 0.5 \\ \cline{2-7} 
 & \begin{tabular}[c]{@{}c@{}}0.78\\ (WRN20x1.7)\end{tabular} & 72.17 & \begin{tabular}[c]{@{}c@{}}0.28\\ (RN20)\end{tabular} & 67.0 & 3 & GSLR \\ \hline\hline
\end{tabular}
\end{table}

\subsection{Results}
\label{secsec_expresult}
We compare our models with the previous works in~\tablename~\ref{table:compare}. The compared QDNN models trained with KD include QDistill~\cite{polino2018model}, Apprentice~\cite{mishra2018apprentice}, and Guided~\cite{zhuang2018towards}. We achieve the results that significantly exceed those of previous studies. We compare our model (0.27M) with the `student model 2' (SM 2) of QDistill that has 0.3 M parameters, and achieve an 18.32\% of performance gap in the test accuracy. Also, it is about 1\% better than the ResNet20 result reported by Apprentice and even achieved the same performance with their ResNet32 result. When quantized to 1-bit, the test accuracy of 91.3\% is obtained, which is almost the same as Apprentice's ResNet20 2-bit model. In the case of CIFAR-100, QDistil and Guided student models use considerably large number, 17.2M and 22.0M, of parameters. Our student model only contains 0.28M parameters but achieve 17.7\% and 2.4\% higher accuracies than QDistill and Guided, respectively. These huge performance gaps show the importance of selecting the proper hyperparameters.

\begin{figure}[t]
    \centering
    \subfigure[CIFAR-10]{\includegraphics[width=0.495\linewidth]{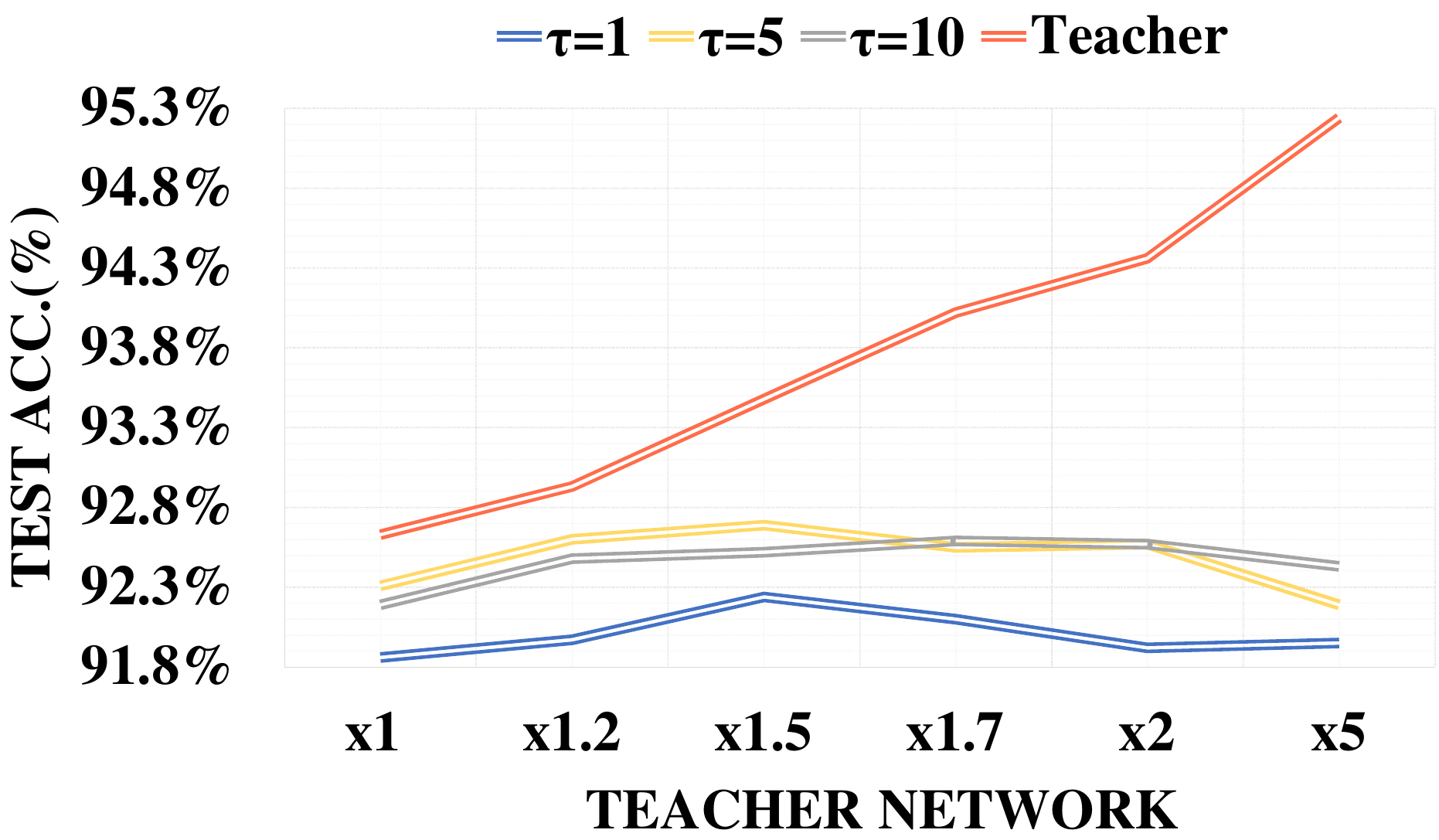}}\hfill
    \subfigure[CIFAR-100]{\includegraphics[width=0.495\linewidth]{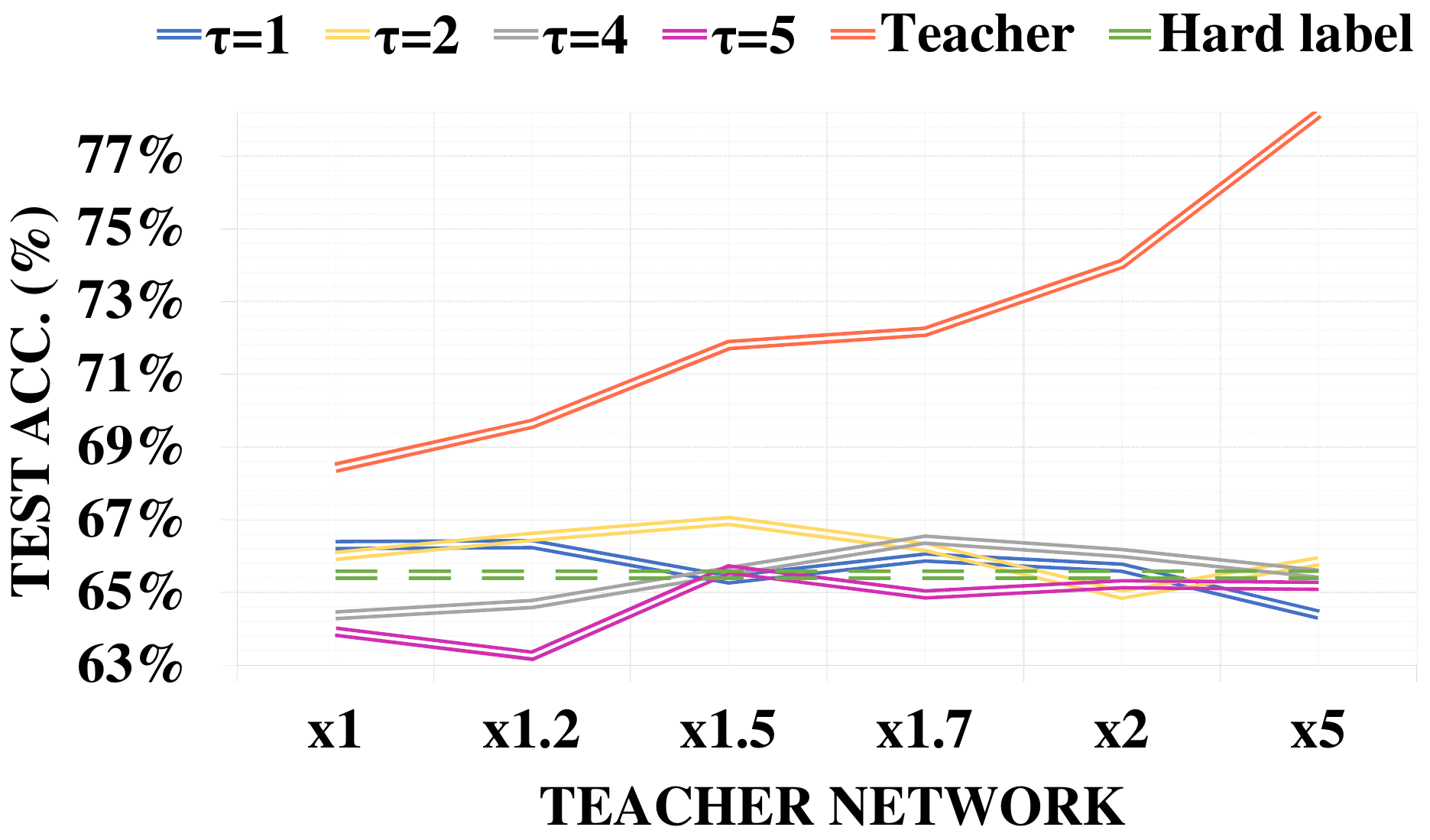}}
 \vspace{-0.5cm}
    \caption{Results of 2-bit ResNet20 that trained with varying the \textit{temperature} ($\tau$) and the \textit{size of the teacher network} on the CIFAR-10 and the CIFAR-100 datasets. The numbers in x-axis represent the wide factor ($N$) for WideResNet20x$N$.}
    \label{fig:TvsModel}
\end{figure}

\subsection{Model size and temperature}
\label{secsec_size_temperature}
We report the test accuracies of 2-bit ResNet20 on CIFAR-10 in \figurename~\ref{fig:TvsModel} (a). To demonstrate the effect of the \textit{temperature} for the QDNN training, we train 2-bit ResNet20 while varying the \textit{size of the teacher network} from `WRN20x1' to `WRN20x5'. Each experiment was conducted for three $\tau$ values of 1, 5, and 10, which correspond to small, medium, and large one, respectively. Note that WRN20x$N$ contains the number of channel maps increased by $N$ times. When the value of $\tau$ is small (blue line in the figure), the performance greatly depends on $N$, or the teacher model size. The performance change is much reduced as the value of $\tau$ increases to the medium (orange line) or the large value (blue line). This is related to the accuracy of the teacher model (red line). When the \textit{size of the teacher model} increases, the shape of the soft label becomes similar to that of hard label. In this case, the KD training results are not much different from that trained with the hard label. Therefore, with $\tau=1$, the performance decreases to 91.9\% when the teacher network becomes larger than WRN20x2. This result is similar to the performance of a 2-bit ResNet20 trained with the hard loss (91.71\%). The soft label needs to have a broad shape and it can be achieved either by increasing the \textit{temperature} or limiting the \textit{size of the teacher network}. A similar problem can occur for full-precision model KD training, but it is more important for QDNN since the model capacity is reduced due to quantization. Therefore, when training QDNN with KD, we need to consider the relationship between the \textit{size of teacher model} and the \textit{temperature}.

\figurename~\ref{fig:TvsModel} (b) shows the test accuracies of the 2-bit ResNet20 trained with KD on the CIFAR-100 dataset. Since the CIFAR-100 includes 100 classes, the soft label distribution is not sharp and the optimum value of $\tau$ is usually lower than that of the CIFAR-10. More specifically, when $\tau$ is larger than 5 (purple line), the test accuracies are lower than 65.49\% (green dotted line), the accuracy of the 2-bit ResNet20 trained with the hard label. The soft label can easily become too flat even with a small $\tau$, thus the teacher's knowledge does not transfer well to the student network. When $\tau$ is not large (e.g. less than 5), the tendency is similar to CIFAR-10 experiment. When $\tau$ is 1 (blue line), the best performance is observed with ResNet20. As $\tau$ increases to 2 (yellow line) and 4 (grey line), the size of the best performing teacher model also changes to WRN20x1.5 and WRN20x1.7, respectively. This demonstrates that a proper value of \textit{temperature} can improve the performance, but it should not be too high since the knowledge from the teacher network can disappear.


\begin{figure}[t]
    \centering
    \subfigure[KD]{\includegraphics[width=0.495\linewidth]{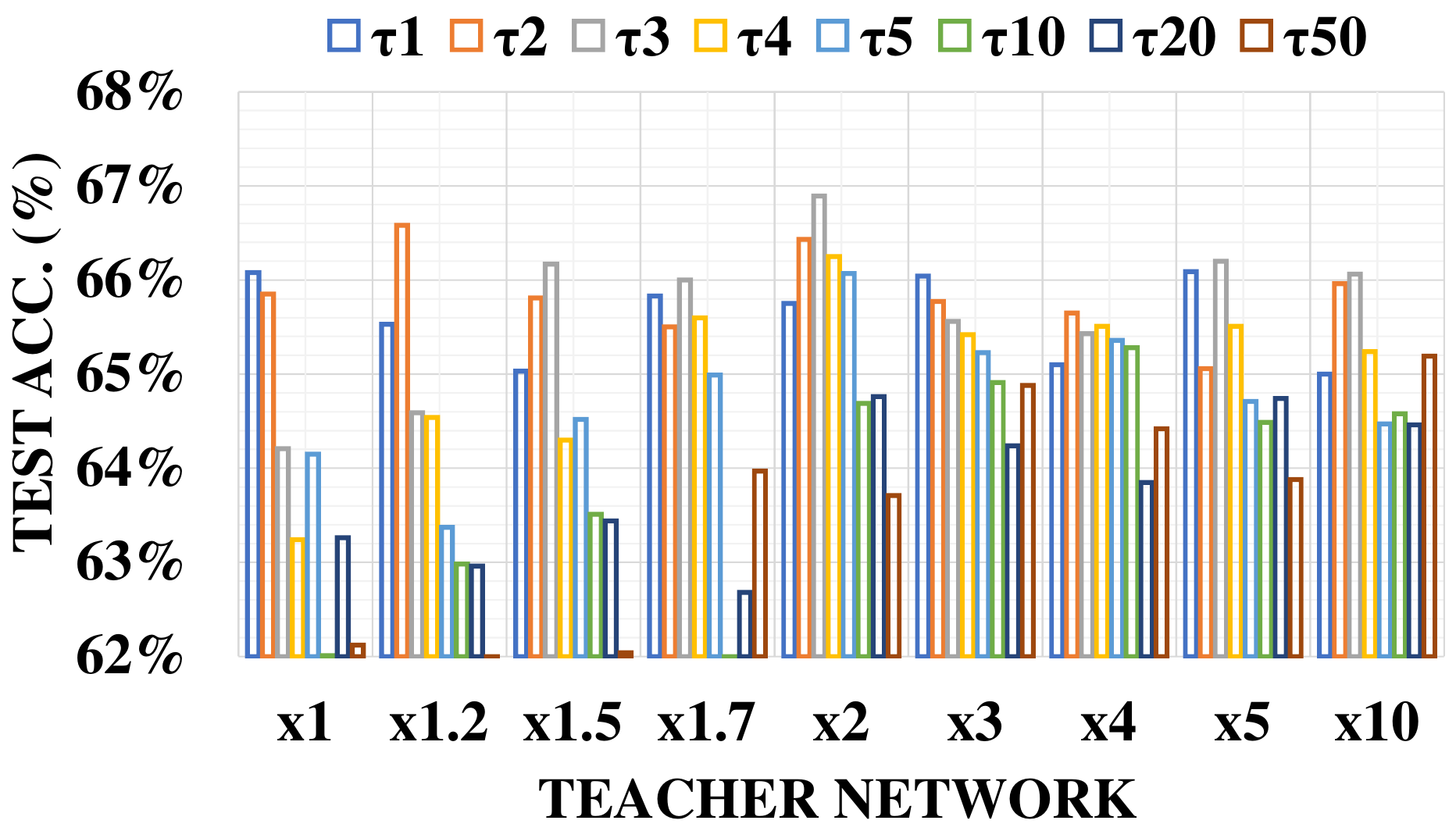}}\hfill
    \subfigure[KD+GSLR]{\includegraphics[width=0.495\linewidth]{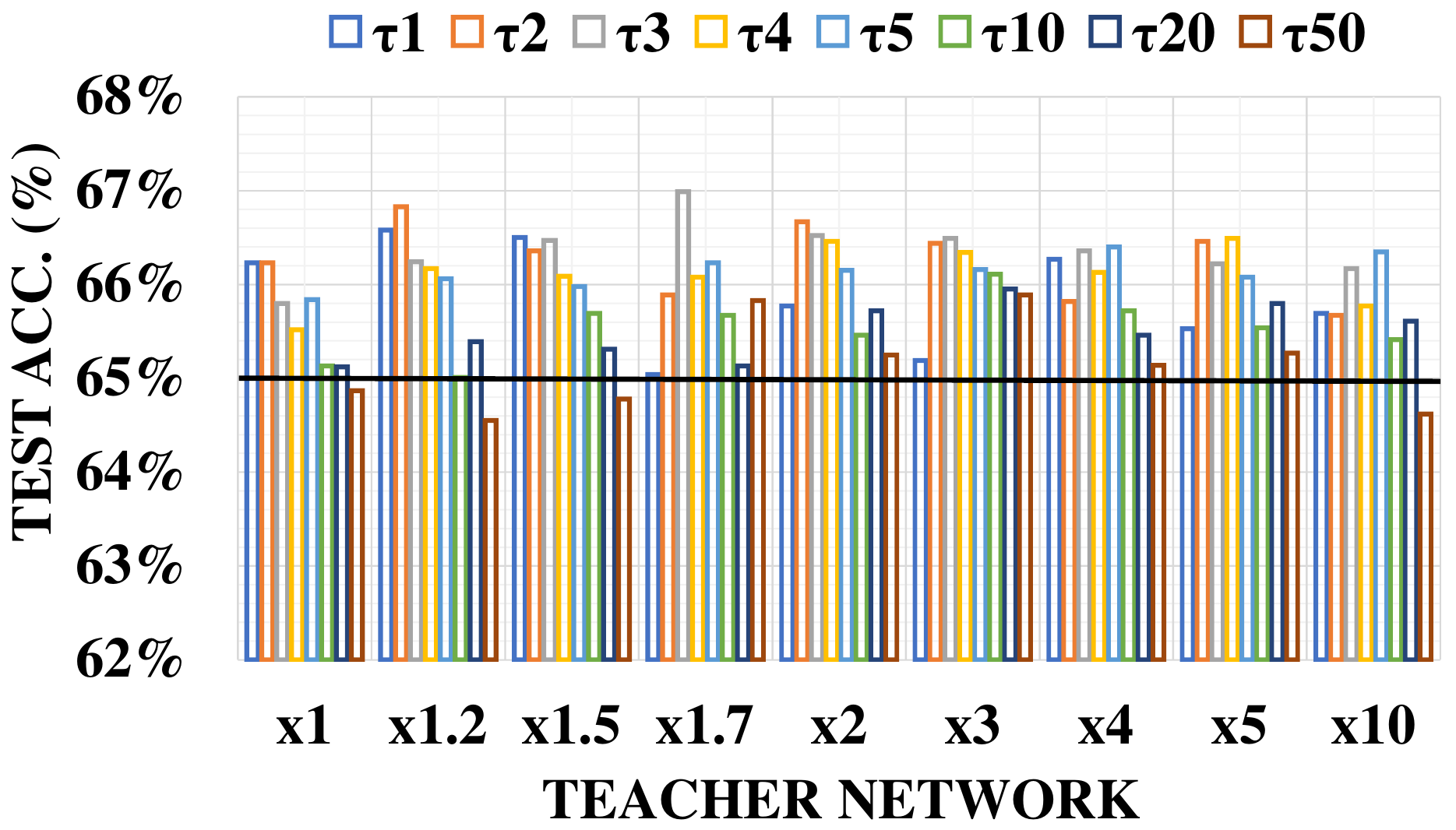}}
 \vspace{-0.5cm}
    \caption{Results of 2-bit ResNet20 models that trained by the various \textit{size of teacher networks} and the \textit{temperature} on CIFAR-100. In (b), the black horizontal line represents the test accuracy when the student network is trained with hard label only.}
    \label{fig:HTKDCR}
\end{figure}

\subsection{Gradual soft loss reducing}
\label{secsec_CR}
Throughout the paper, we have discussed the effects of the \textit{temperature} and the \textit{size of teacher network} on the QDNN training with KD. Since the two hyperparameters are interrelated, careful parameter selection is required and it makes the training challenging. We also have the risk of cherry picking if the outcome cannot be predicted well without using the test result. Thus, we need to have a parameter setting technique that is fail-proof. 

We have developed a KD technique that is much less sensitive to specific parameter setting for KD. At the beginning of the training, where the gradient changes a lot, we use the soft and hard losses equally and then, gradually reduce the amount of the soft loss as the training proceeds. We name this simple method as the \textit{gradual soft loss reducing} (GSLR) technique. To evaluate the effectiveness of the GSLR, we train 2-bit ResNet20 while varying the \textit{size of the teacher} and the \textit{temperature} as shown in \figurename~\ref{fig:HTKDCR}. The results clearly show that GSLR greatly aids to improve the performance or at least yields the comparable results with the hard loss (black horizontal line). When comparing the traditional KD, shown in \figurename~\ref{fig:HTKDCR} (a), and GSLR KD, in \figurename~\ref{fig:HTKDCR} (b), we can find that the latter yields much more predictable result, by which reducing the risk of cherry picking.

\section{Concluding Remarks}
\label{sec_conclude}
In this work, we investigate the teacher model choice and the impact of the hyperparameters in quantized deep neural networks training with knowledge distillation. We found that the teacher needs not be a quantized neural network. Instead, hyperparameters that control the shape of softmax distribution is more important. The hyperparameters for KD, which are the \textit{temperature}, \textit{loss weighting factor}, and \textit{size of the teacher network}, are closely interrelated. When the \textit{size of the teacher network} grows, increasing the \textit{temperature} aids to boost the performance to some extent. We introduce a simple training technique, \textit{gradual soft loss reducing} (GSLR) for fail-safe KD training. At the beginning of the training, GSLR equally employs the hard and soft losses, and then gradually reduces the soft loss as the training proceeds. With careful hyperparameter selection and the GSLR technique, we achieve the far better performances than those of previous studies for designing 2-bit quantized deep neural networks on the CIFAR-10 and CIFAR-100 datasets.

\bibliographystyle{IEEEbib}
\bibliography{refs}

\end{document}